\documentclass{article}

\usepackage{arxiv}

\usepackage[utf8]{inputenc} 
\usepackage[T1]{fontenc}    
\usepackage{hyperref}       
\usepackage{url}            
\usepackage{booktabs}       
\usepackage{amsfonts}       
\usepackage{nicefrac}       
\usepackage{microtype}      
\usepackage{lipsum}
\usepackage{graphicx}
\usepackage{authblk}
\graphicspath{ {./images/} }

\usepackage{subcaption}     

\title{Complex data labeling with deep learning methods: Lessons from fisheries acoustics}

\author[1,2]{J.M.A.Sarr} 
 \author[1,2]{T. Brochier}
 \author[3,4]{P.Brehmer} 
 \author[3]{Y.Perrot} 
 \author[1,2]{A.Bah} 
 \author[4]{A.Sarré} 
 \author[5]{M.A.Jeyid} 
 \author[6]{M.Sidibeh}
 \author[7]{S.El Ayoub}
  \affil[1]{Université Cheikh Anta Diop de Dakar UCAD, Ecole Supérieure Polytechnique, BP 15915, Dakar Fann, Senegal}
  \affil[2]{IRD, Sorbonne Université, UMMISCO, F-93143, Bondy, France}
  \affil[3]{IRD, Univ Brest, CNRS, Ifremer, LEMAR, Plouzané, France}
  \affil[4]{ISRA/CRODT, Pole de recherche de Hann, BP2241, Dakar, Senegal}
  \affil[5]{IMROP, BP22, Nouadhibou, Mauritania}
  \affil[6]{Fisheries Department (FD), Marina Bay, Banjul, The Gambia}
  \affil[7]{INRH, Anza 80000, Agadir, Morocco}
  
\begin{document}
\maketitle
\begin{abstract}
Quantitative and qualitative analysis of acoustic backscattered signals from the seabed bottom to the sea surface is used worldwide for fish stocks assessment and marine ecosystem monitoring. Huge amounts of raw data are collected yet require tedious expert labeling. This paper focuses on a case study where the ground truth labels are non-obvious: echograms labeling, which is time-consuming and critical for the quality of fisheries and ecological analysis. We investigate how these tasks can benefit from supervised learning algorithms and demonstrate that convolutional neural networks trained with non-stationary datasets can be used to stress parts of a new dataset needing human expert correction. Further development of this approach paves the way toward a standardization of the labeling process in fisheries acoustics and is a good case study for non-obvious data labeling processes.
\end{abstract}


\section{Introduction}
\subsection{Cost and reliability of the labeling processes in data driven applications: the case of fisheries acoustics}

Solving pattern recognition problems with machine learning algorithms strongly rely on the  availability of reliable ground truth datasets. For obvious labeling tasks as naming pictures of  everyday life objects, large reliable ground truth datasets can be acquired at relatively low cost,  e.g. using crowd labeling methods (Griffin et al. 2007; Deng et al. 2009). In the case of more  complex data, the labeling must be done by experts, which increases costs but also brings  incertitude on finding ground truth labels. Indeed, for some complex data, different experts may  find different labels. This issue is particularly experienced in medical computer-aid diagnosis  (Raykar et al. 2010) but also in remote sensing applications, e.g: in object detection with radar  (Smyth et al. 1995) or in fisheries acoustics (McClatchie et al. 2000).  

Fisheries acoustics is the main nondestructive method for estimating the abundance of  pelagic and semi-pelagic fish (Simmonds and MacLennan 2005; Brehmer 2006). These  estimations are critical for fisheries worldwide because they allow managers to provide  recommendations on fishing effort level adjustment to avoid overexploitation, maintain  ecosystem health, and ensure food security. Large datasets have been routinely collected  worldwide to estimate fish abundance (MacLennan 1986) or, for example, used to study fish  behavior (Guillard et al. 2010), and even in aquaculture (Brehmer et al. 2006). Labeling these  datasets is a keypoint for any further quantitative or qualitative analysis. In this paper, we  focused on the expert labeling process of these data.  

One of the first operations needed before any inference for assessing fish abundance can  be drawn is to identify the bottom depth along the survey path (Korneliussen 2004; MacLennan et al. 2004). Indeed, accurate correction of the bottom line is needed because fish abundance is  estimated by integrating the acoustic signal from bottom to surface. Since the bottom acoustic  signal is often strong, small errors in bottom depth estimation can lead to the overestimation of  fish abundance in the water column. Indeed, if the bottom depth estimated is miscalculated and  falls below the true bottom depth, the echo integration may contribute substantial errors because  it will interpret the amount of energy backscattered by the bottom as biological resources (Ona  and Mitson 1996; Villalobos et al. 2013).

This paper investigates how supervised learning can help automate the labeling process  of the bottom correction as a case study. Nowadays, bottom-detection algorithms with a single beam echo sounder rely on echo-amplitude measurements within a depths range specified by an  onboard operator as the upper and lower depth limits most likely to be used during an acoustic  sea survey (e.g., see “bottom detection” in Simrad EK60). However, this procedure can fail for a  variety of reasons, e.g., either coming from onboard errors in manual setting of the instrument, or  from noise in the reflected signal itself that can perturb the bottom-detection algorithm. For  example, over soft and weakly reflecting grounds, the bottom may be detected below its true  level. Also, a high density of fish present near the seabed can generate a false detection of the  bottom echo (MacLennan et al. 2004). Thus, before producing fish stock assessments from these  observations, the signal must be hand post-processed by an expert to find and correct these errors  (Socha et al. 1996). This consists of visually scrutinizing the entire echogram for bad data  sections and poor bottom detection and then removing questionable data and redefining the  bottom as needed (Bartholomä 2006). This task is expensive because it requires an expert to go  through all of the echograms, i.e.several millions of pings, depending on the cruise duration. This is a typical case of a labeling process where the ground truth label is non-obvious and varies  from one expert to another. 

Surprisingly, few studies in the fisheries acoustics field have focused on improving  bottom detection or correction (Foote et al. 1991), and most have concentrated on the  discrimination of sediment or seabed classification (e.g., Bartholomä 2006) and obviously  biological targets identification (Simmonds and MacLennan 2005; Brehmer et al. 2019;  Brautaset et al., 2020).  

\subsection{Machine learning to automate the data labeling process?}
The problem at hand is typical of the challenges faced by the remote sensing (RS) community.  Indeed RS data are expensive to collect, error-prone, and require expert interpretation. Ball et al. (2017) suggested machine learning (ML) and deep learning (DL) methods to automate the  human-engineering process. Fisheries acoustics data (provided as an echogram matrix) are  inherently highly dimensional. Furthermore, as underwater feature extraction methods gradually  become less effective with the expansion of acoustic datasets, the need to find automated  methods and procedures to extract meaningful features is sharper.  

Deep Neural Networks (DNN) are neural networks with multiple hidden layers. They can  learn a representation of the data with multiple levels of abstraction (hidden layers) and are well  suited to making inferences from large volumes of complex data in an end-to-end fashion (Lecun et al. 2015). The premise of DL is to partially automate the feature engineering made by humans  as they often suffer from biases. Indeed, it is often the case that two experts might not agree on  the correction of a given echogram.

However, only a few attempts to apply DL and ML were achieved in fisheries acoustics.  We cite Williams (2016) and Denos et al. (2017) who first attempted to classify underwater detected objects with DL, but were facing the problem of insufficient availability of training  data. Recently, Brautaset et al., (2020) used a convolutional autoencoder architecture for acoustic  fish school detection and classification, and got promising results but still facing the problem of  the training dataset quality. In the case of fish school identification, overpassing the problem of  training dataset quality is complicated because of the difficulty to get a “field truth” of which  species reflected the signal (Simmonds et al. 2008).  

Two approaches are possible using ML methods: (1) to directly predict the bottom value  from the echogram and (2) to evaluate the quality of the pings by classifying them into two  groups, depending on whether the bottom needs correction or not. The first approach could be to  develop a system to fully automate human intervention, but errors would be harder to spot.  Indeed, when the biological resource being targeted for echointegration require high precision,  human intervention would still be needed, and the first automation procedure would not give any  insight to the expert to spot them. So, we took a step back and decided to address the problem of  reducing the expert’s time to perform the most basic, but also the most time-consuming dataset  post-processing task needed for direct fish stock assessment, i.e. the bottom depth correction. We  designed a system that would help the expert to gain time by highlighting pings with a higher  likelihood of requiring correction. Another benefit of this method is that by automating the  labeling process, it removes the inherent variability among different experts and makes the  labeling process unified for different sea surveys. Here the idea is to lay down a methodology  that would leverage datasets collected in past campaigns to support the bottom correction process  on a newly collected dataset. Hence, the ML task is to provide a “quality label” for each part of the dataset so that the expert can focus only on the echogram sections that are likely to need  correction, i.e., to avoid replaying the whole survey. 

\subsection{Contributions}
We evaluated several ML procedures to leverage past labeled data for which an expert has  already corrected the bottom line prediction, thus large training datasets are available. The  specific task investigated was to identify the parts of the echogram requiring the expert to correct  the bottom line. Our first contribution is to propose a comparison of different machine learning  algorithms for this task. Four learning algorithms are compared: Random Forests, Convolutional  Neural Networks, Support Vector Machines, and Feed-Forward Neural Networks. Their  hyperparameters were found using Bayesian Optimisation. This led to the identification of CNN  as the most adapted learning algorithm.  

In most applications when the best ML algorithm is discovered, it is used in production  on new datasets. However, in our case, a new dataset may have different attributes such as a shift  in the label distribution, or a different noise structure. For instance, the frequency of pings  requiring an expert correction might be different from one sea campaign to another as discussed  above. Also, the noise might not be distributed equally from one sea campaign to another. This  has motivated our second contribution, we provide experimental evidence showing the beneficial  effect of mixing training datasets from different sea surveys to improve performance at test time;  we refer to this technique as cross-domain training in the following.

\begin{figure}[h!]
	\centering
	\includegraphics[scale = 0.6]{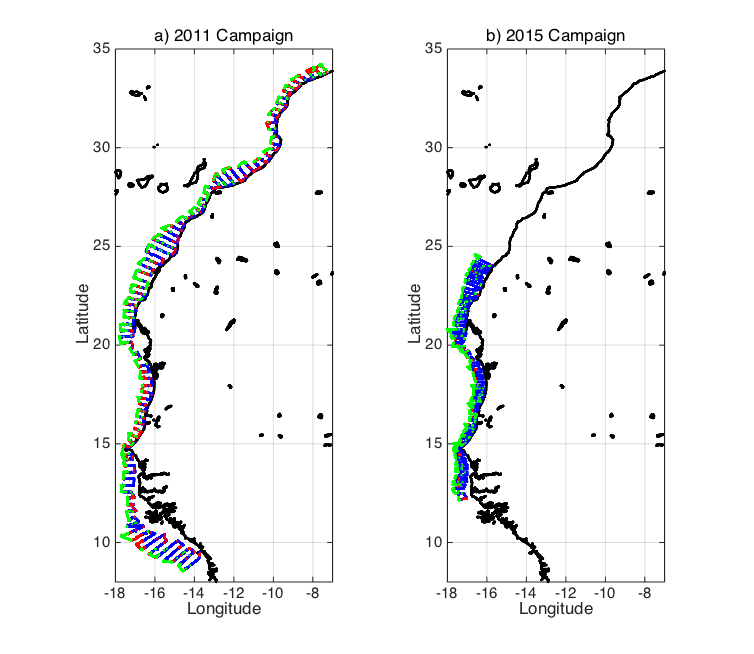}
	\caption{Survey design carried out off northwest Africa during acoustic annual assessment surveys at sea in 2011 and 2015 (Research Vessel Dr. Fridtjof Nansen). The classes appear in the following order: in green, the part of the echogram with no bottom; in blue, the part of the echogram requiring only a weak correction from the expert; in red, the pings requiring a strong correction from the fisheries acoustic experts. . (For interpretation of the references to color in this figure legend, the reader is referred to the web version of this article.) \label{fig:map}}
\end{figure}

\begin{table}[h!]
\centering
\caption{Summary of the variables present in each 2011 and 2015 sea survey datasets dimensions and associated labels. The main variables used to process the data and furthermore to set up the machine learning model are described.
$S_{v}$ means volume backscattering strength (in dB).}
    \begin{tabular}{|p{3cm}|p{3cm}|p{3cm}|p{3cm}|p{3cm}|}
    \hline
    & Depth & Echogram & Bottom & Clear Bottom \\
    \hline
    2015 dataset dimensions: [Nb. rows, Nb. columns] & [2567, 1] & [2567, 2 000 000] & [1, 2 000 000] & [1, 2 000 000] \\
    \hline
    2011 dataset dimensions: [Nb rows, Nb columns] & [2581, 1] & [2581, 2 661 003] & [1, 2 661 003] & [1, 2 661 003] \\
    \hline
    Units & Meter (m) & $S_{v}$ in dB & Meter (m) & Meter (m) \\
    Horizontal/vertical vectors & Vertical vector: every cell correspond to a value in meter & Matrix: every column is a given ping, and every row a value in dB corresponding to a depth & Horizontal vector: every column gives the bottom depth found by the automatic procedure & Horizontal vector: every column gives the bottom depth corrected by the expert \\
    \hline
    \end{tabular}%
\end{table}

\section{Materials and Methods}
In section 2.1 we provide an exhaustive description of the crude dataset collection, formatting,  labeling method, and the subdivision in training, validation, and test datasets. Computing means  and associated limitations on the size of the processed dataset are also described here. In section  2.2 we present the methodology for learning algorithms comparison, for training dataset  optimization and the training settings. 

\subsection{Data Processing}
\subsubsection{Description of the crude dataset }
Acoustic data came from an international collaboration of northwest African fishery research  centers, which gathered their data at the subregional level. The data consist of two datasets  corresponding to two campaigns from the Nansen project (Fisheries Research Vessel Dr. Fridtjof  Nansen) that took place in 2011 and 2015 off northwest Africa (Fig 1) (e.g., Sarré et al. 2018).  Here, each acoustic pulse is called a ping, and we call the matrix obtained from gathering the  backscattered signals of a sequence of pings the echogram. These echograms usually come from  preprocessed acoustic surveys at sea (Fig 1). The data preprocessing was done using the Matecho  software (Perrot et al. 2018). The backscattered signal from the upper 500 m of the water column  was extracted, and the echo at each depth was interpolated on a regular grid with a vertical  resolution of 20 cm. Still, during preprocessing, the ocean depth for each ping was estimated by  an automatic algorithm that searched for the maximum gradient of the acoustic signal, usually  corresponding to the acoustic signal at the bottom. Then in post-processing, the echogram and the automatic bottom line detection were visualized by an expert who manually performed the  correction of the bottom line. 

Data were acquired with an echosounder fixed to the hull of the vessel. The vessel sent  out acoustic wave pulses of four distinct frequencies in the water at a pulse length of 1 ms. In this  study, we used the 38 kHz frequency as it is a common frequency used in fisheries acoustics and  is not limited over the continental shelf (500–600 m maximum depth). Table 1 gives a summary  of the variables present in each dataset used. The crude echogram presents some irregularities  due to the onboard recording settings. Typical errors of the automatic procedure for bottom line  detection are shown in Fig. 2a., and it can be seen that the expert (red line) roughly cut this part  to avoid including the bottom signal in the echo-integration.  

In particular, NaN (not a number) values were usually present between 500 m (the  maximum recording depth in this case study) and $\sim$20–30 m below the predicted bottom. This is  because, during data collection at sea, the echo sounder operator(s) set the maximum depth to  limit data acquisition to the water column. Nevertheless, in some cases, real values were still  attributed to much greater depths under the actual bottom (Fig 2). Furthermore, these  irregularities were distributed differently between the 2011 and 2015 datasets (Fig 2) as they  were dependent on the survey configuration at sea vs. the local depth surveyed. These  irregularities, common in fisheries acoustics sea surveys, typically challenge traditional ML  approaches. Indeed, to deal with such differences in the dataset’s noise the modeler is often  required to build complex pipelines for feature extraction. The rationale behind the use of DL is  to learn in an end-to-end fashion, with as few preprocessing operations as possible.

\begin{figure}[h!]
	\centering
	\includegraphics[scale = 0.15]{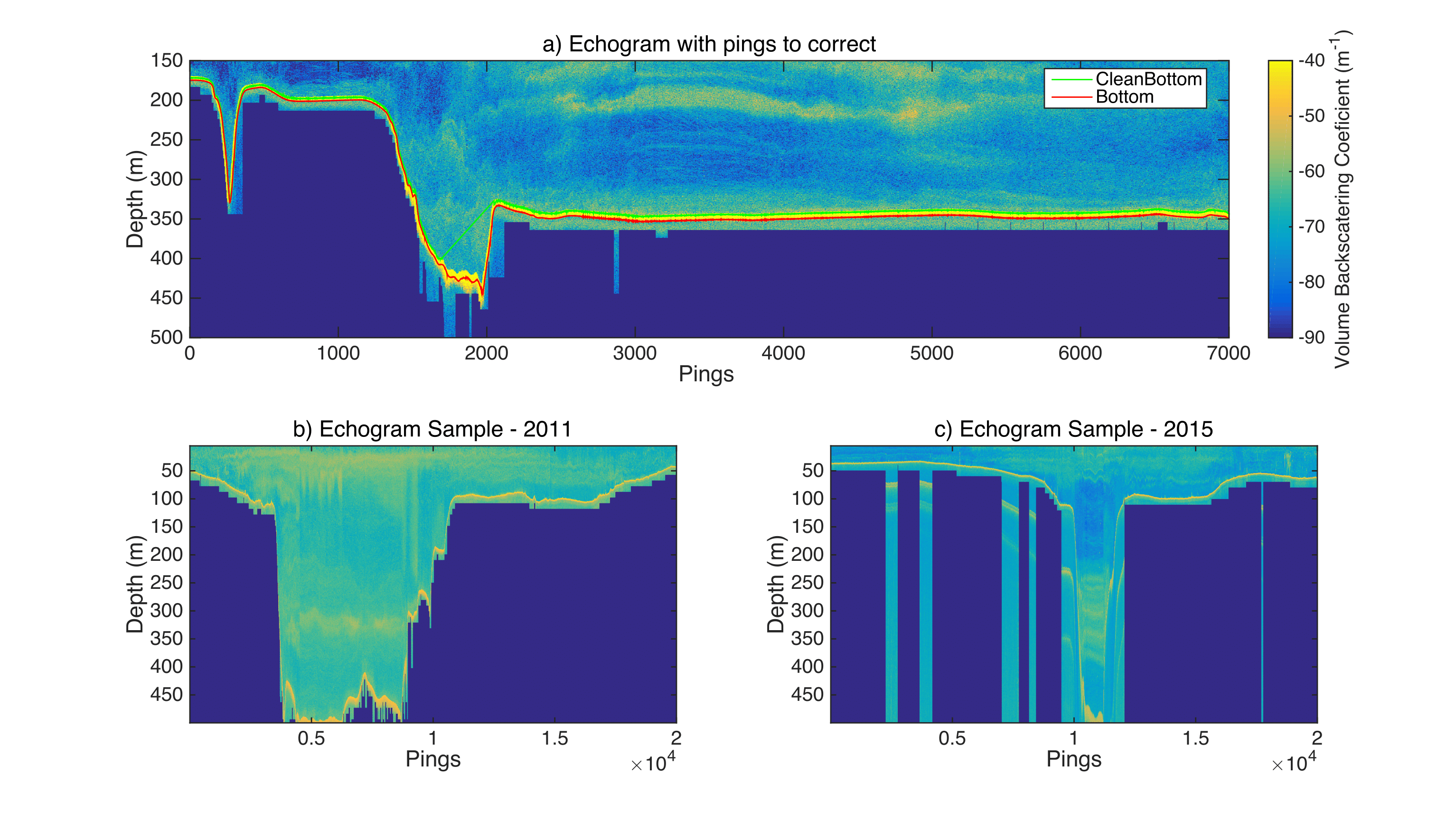}
	\caption{Echogram extracted from the acoustic sea survey used in the study. (a) An example in which the automatic procedure for bottom detection (green line) has failed around ping 2000. (b) Random samples from the 2011 and (c) 2015 echograms, with the same ping size and cell number; showing differences with respect to the settings for NaNs (“not a number”, strong blue color) below the bottom. Unit: see color bar in panel (a). . (For interpretation of the references to color in this figure legend, the reader is referred to the web version of this article.) \label{fig:echo}}
\end{figure}

\begin{figure}[h!]
	\centering
	\includegraphics[scale = 0.45]{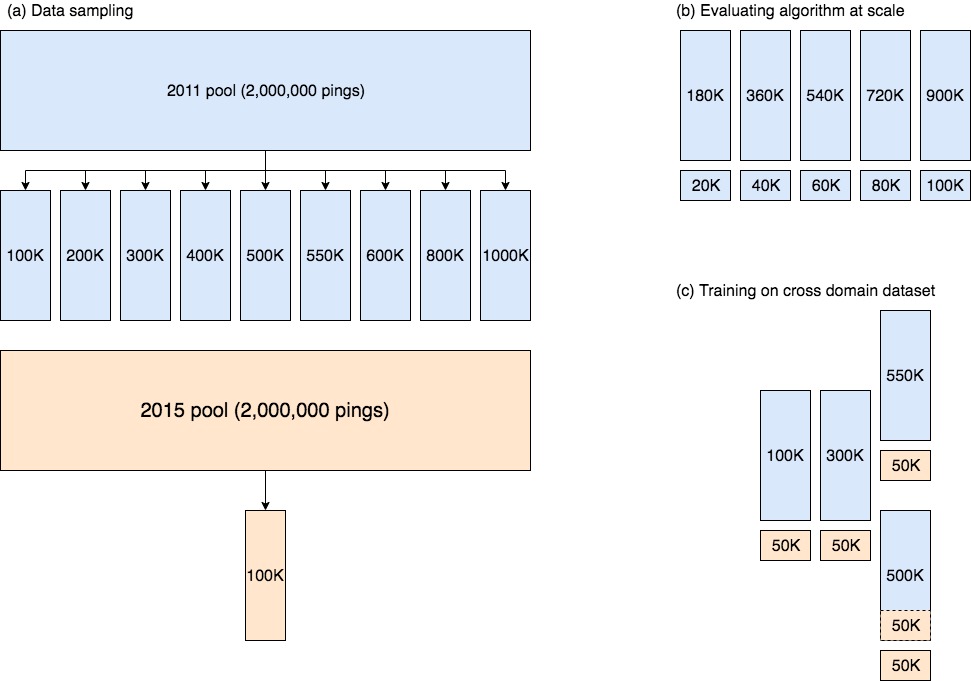}
	\caption{Sampling methodology. Subfigure (a) summarizes the data sampling strategy. Subfigure (b) shows the training and test set sizes as described in experiment 2.2.1. Subfigure (c) shows the setting proposed for experiment 2.2.3. The blue color corresponds to data from 2011 and the orange color corresponds to data from 2015. In (b) and (c), the longer boxes correspond to the training dataset. In (b) the small boxes are used as a test set and in (c) as a validation set. \label{fig:sampling}}
\end{figure}

\begin{table}[h!]
\centering
\caption{2011 and 2015 sea survey dataset after removing pings without bottom and after removing the first cells of each dataset so that they share the same number of rows.}
    \begin{tabular}{|c|c|c|}
    \hline
    & Echogram & Label \\
    \hline
    2015 formatted dataset: [No rows, No columns] & [2550, 1 851 950] & [1, 1 851 950] \\
    \hline
    2011 formatted dataset: [No rows, No columns] &
    [2550, 2 321 967] & [1, 2 321 967] \\
    \hline
    \end{tabular}%
\end{table}

\subsubsection{Formatting the dataset }
Three transformations were applied to extract a standardized format and minimalist training  dataset from the crude one. We reduced the size of the training dataset to reduce learning and  upload times on remote calculation servers. 

The first transformation was carried out to get the same number of rows for our datasets,  which initially varied slightly (Table 1). To do that, we removed the first 31 rows from the entire  2011 echogram ($\sim$ 6 meters), as well as the first 17 rows of the 2015 echogram ($\sim$ 3.3 meters).  This was required because frequently there were strong signals in the first rows that subsequently  disturbed the learning process. Such a transformation was necessary to allow the same neural  networks to be trained and applied to the two dataset subsamples. Finally, we were left with 2550  rows in each of the preprocessed datasets. 

The second transformation was aimed at reducing the size of the dataset by removing  pings for which no learning could be extracted, to distinguish between bottom qualities, i.e.,  when pings in the echogram did not hit the bottom. This occurred when the vessel was located in  areas deeper than 500 m because only the first 500 m were recorded during the sea campaign.  These pings were removed using a threshold-based filter. We discriminated between the  echograms with and without a bottom, based on the presence or absence of strong backscattering  (> $-$32 dB), considered as a bottom signature. Finally, as we needed every ping to have the same  dimensions for further calculation, and numerical algorithms could not process NaN values, we  replaced all of the NaN values by -200 dB, a value that corresponds to the weakest recorded  values in the echograms (-199 dB for 2011 and -198 dB for 2015). This operation can be seen  as replacing non-recorded values by noise. Indeed, values below $\sim-$90 dB are never (or seldom)  considered in fisheries acoustics analysis. Thus, NaN values were treated as area-located under the bottom and not reached by the acoustic signal. The sizes of the 2011 and 2015 formatted  datasets are summarized in Table 2. Finally, both datasets were standardized before training. 

\subsubsection{Data labeling}
We followed a generic ML methodology (Goodfellow et al. 2016). The goal of the procedure is  to classify the data (here, acoustic pings) into two classes before the intervention of the expert to  rationalize their work when correcting the crude echogram. The first class gathers the pings on  which the automatic procedure usually performed well, thus needing less attention from the  expert. The second class regroups the pings for which the expert usually needed to correct the  automatic bottom prediction. We labeled each remaining ping (after formatting; see the previous  section) as belonging to one of these two classes according to the distance between the expert  correction (variable "CleanBottom", Table 1) and the initially predicted bottom depth (variable  “Bottom,” Table 1). Thus, if |CleanBottom - Bottom| < 3.31, pings were labeled “weak  correction,” i.e., no or weak expert correction needed, while if |CleanBottom - Bottom| >= 3.31,  pings were labeled “strong correction,” i.e., major expert correction needed. Examples of these  two classes are shown in Fig 2. The threshold value of 3.31 was chosen by comparing the  accuracy obtained after one epoch of training for different threshold values ranging from 1.00 to  5.00 with a step of 0.01, and we selected the threshold that gave us the best classification  accuracy. 
The distribution of those classes varied significantly from 2011 to 2015. Indeed, the class  of strong correction accounted for 13\% of the pings in 2011, whereas it accounted for only 1\% in  2015. As can be seen in Fig 1, there is no clear pattern from the pings to correct. Hence, we  expect to use ML methods to automate the finding of the ping, with a high likelihood of  requiring a strong correction.

\subsubsection{Sampling methodology }
Here we describe datasets that were used to (1) optimize hyperparameters, (2) compare learning  algorithms performances, and (3) train the best-identified learning algorithm with a mixed  dataset. To begin with, the first 2,000,000 pings of each campaign were selected to constitute the  pool of data. 2,000,000 was the maximum we can fit into memory. In the following, to ease the  notation every 1000 pings are going to be treated as 1K pings. 
For the Bayesian optimization, we used a dataset of 100K pings randomly extracted from  the complete 2011 pool dataset (Fig 3a). For comparing learning algorithms performances,  datasets with successive sizes of 200K, 400K, 600K, 800K, and 1000K pings were also extracted  from the 2011 pool dataset (respectively 2.0, 4, 6.0, 8.0, and 10.0 Go). The largest dataset used  for training was made of 1000K pings because it was tedious to upload large datasets on GPU  clusters online. Furthermore, to compare different learning algorithms, each of these datasets was  split into 90\% of the pings for the training set and 10\% for the test set (Fig 3b). To evaluate the  effect of cross-domain training on the best-identified algorithm, 100K, 300K, 500K, and 550K  pings were sampled from the 2011 pool dataset for training along with 100K pings following  each other from the 2015 pool. Those 100K pings were further randomly divided into two  datasets of 50K pings, one that would serve as a validation dataset and one that would serve for  cross-training (Fig 3c).  
\subsubsection{Computing Means and Limitations }
Formatting of the dataset and the labeling operations were performed on a personal computer.  For training purposes, we uploaded the dataset on a cloud platform to perform further calculations with graphical processing units (GPUs). However, the local Internet connection  speed and stability limited the size of the dataset that could be uploaded. Also, the size of the  unzipped 2011 preprocessed echogram was 22.81 Gigabytes, which scarcely fitted into random  access memory (RAM). For training algorithms, a maximum of 1000K pings has been used to  allow the dataset to be uploaded.  

\subsection{ Experimental set up }
\subsubsection{Methods for Comparing learning algorithms performances }
The first experiment was to compare the ability of different algorithms to learn from  echosounder datasets at different scales. The compared algorithms suggested in the literature,  were Random Forests (RF), Support Vector Machines (SVM), Feed-Forward Neural Networks  (FFNN) and Convolutional Neural Networks (CNN) (LeCun and Bengio 1995; Niu et al. 2017 ; Ferguson et al. 2018). 

A principled approach to compare machine-learning algorithms is to find their respective  hyperparameters following a single optimization method. Here we used Bayesian Optimization  (BO) as opposed to grid search and random search, as it has been shown that Bayesian  optimization finds better hyperparameters significantly faster than random search which is  superior to grid search (Snoek et al. 2012). Moreover, BO surpasses a human expert at selecting  hyperparameters on the competitive CIFAR-10 dataset frequently used to benchmark computer  vision algorithms (Snoek et al. 2012). 

In this study, we used the open source python Library GyOpt (SheffieldML, 2020) for BO, with expected  improvement as acquisition function and Gaussian Processes to model the surrogate function  with the acquisition parameter equal to 0.1. For each algorithm, we fixed the range of variation  for a set of hyperparameters. For RF the most sensitive hyperparameters according to (Niu et al. 2017) are the number of decision trees (range [10,1000]) and the minimum samples per leaf  (range [20,50]). Concerning SVM we chose a linear kernel as the number of features is important  (2500) (Hsu et al. 2003) and we optimize only the hyperparameter C (range [10, 1000]). The  FFNN was chosen to have 3 hidden layers (ranging respectively [5,600], [5, 320] and [5,120]  neurons) and with dropout (Srivastava et al. 2014) for the last hidden layer (ranging [0,1]. Every hidden layer used  SELU (Klambauer et al. 2017) as an activation function. Finally, the CNN hyperparameters were three convolutional  layers (with kernels ranging in [5, 60]) featuring batch normalization (Ioffe and Szegedy, 2015), followed by three hidden layers (with the same range as the  FFNN), and with only dropout applied at the last hidden layer (range [0,1]). For BO, each  learning algorithm was trained with a subset of 100,000 pings from the 2001 dataset, (Fig 3). The  procedure was repeated for 50 iterations. At each iteration, the hyperparameters are tuned and the  best validation accuracy is recorded. The maximum number of iterations was set to 50, but BO  can stop before if hyperparameters converge. 

Once the set of hyperparameters that maximize learning were found for each algorithm,  we trained each algorithm on 5 datasets of sizes: 200K pings, 400K pings, 600K pings, 800K  pings and 1,000K pings ( 2.0, 4, 6.0, 8.0 and 10.0 Go, respectively) to evaluate their respective  ability to scale to big data. Also, the experiment was repeated 5 times to evaluate parameters‘  sensitivity to the learning process. Furthermore, SVM, FFNN, and CNN were trained for 100  epochs each. The test accuracy of both the CNN and the FFNN were obtained using Monte Carlo  Dropout (Gal et al. 2015). In other words, their performance was assessed on the test set 50 times  and the mean value was recorded. Monte Carlo Dropout allows neural networks to express  uncertainty for their prediction due to the dropout factor at the end of the network. It has been  proven to yield better results than standard test evaluation (standalone test run) (Gal et al. 2015).  The SVM and RF were trained with Scikit-learn (Pedregosa et al. 2011). The framework allowed us to use multiprocessing, and the computation was distributed on 6 processors. SVM was  trained with stochastic gradient descent. FFNN and CNN were trained with TensorFlow (Abadi  et al. 2016). Parameters were updated after each epoch using the Adam optimization procedure  with default parameters (Kingma et al. 2014) and binary cross-entropy as cost function. 

\subsubsection{Comparing simple and cross domain training}
The requirement to explore the effect of learning on a cross-domain dataset was to display the  normal learning curve when we scaled the size of the training set and compare it with a learning  curve made with a cross-domain dataset. We compared the effect of mixing during training on  the dataset coming from each pool of data (2011 and 2015). To this end, the subsample of  100,000 pings from 2015 was randomly divided into two datasets of 50,000 pings; one for the  validation set, and the other that would be used to mix with data from 2011 (see Fig 3 and §  hereafter). 

Firstly, we successively trained on two datasets of sizes: 100,000 and 300,000 pings to  get a baseline learning curve, denoted respectively as ST-100K and ST-300K (simple training  100K pings and simple training 300K pings). Secondly, to evaluate the impact of mixing two  datasets with different label distribution and different noise structure we trained on two datasets  of size 550,000 pings described as follows: (1) 550,000 pings randomly sampled from the 2011  pool dataset only (hereafter referred as ST-550) and (2) a mix of 500,000 pings randomly  sampled from the 2011 pool dataset and 50,000 pings randomly selected from the 100,000  subsample ping 2015 datasets (hereafter referred as CDT-550K). The rationale behind the second  approach was to avoid overlearning from the irregularities and data distributions specific to the  2011 dataset, and to assess the effect of mixing during training. Finally, the generalization
performance of each model was evaluated locally by testing the classification accuracy of the  models trained in each setting (ST-100K, ST-300K, ST-550K, and CDT-550K) on two test sets:  on one hand the remaining unseen part of the 2011 dataset, on the other hand, the unseen part of  the 2015 dataset. 

\section{Results}
\subsection{Comparing learning algorithms’ performances}
The best validation accuracy obtained for a set of hyperparameters were obtained in less than 10  iterations for SVM and in about 22 iterations for RF. This is sound, indeed RF has two  hyperparameters and SVM only one. On the other hand, FFNN and CNN improved but were not  settled even after 50 iterations (Fig 4). The final set of hyperparameters for each learning  algorithm is displayed in Table IV. 
Learning algorithms' performances were then compared when trained with increasing size  of datasets. Support Vector Machines (SVM) did not increase its performance when augmenting  dataset size; surprisingly it seems to even worsen (Fig 5a). Feed-Forward Neural Network  (FFNN) exhibits the greater variability, for every size of the training set the worst value is almost  approximately the same and is the result of sticking on a local minimum as the training loss did  stagnate during training. Note that in the experiment on the 800K dataset the algorithm did not  stick in any local minima (Fig 5a). As a result, the mean test accuracy suffered and explained the  drop in performance for the last dataset. Random Forest (RF) displayed almost no variability and  its performance benefited as a result of increasing dataset size. 

Finally, Convolutional Neural Networks (CNN) got the highest mean test accuracy score  right when training on 400K pings and more (Fig 5a). It is furthermore more stable to train as it  displays less variability than FFNN while always having the best maximum test accuracy score.  Except for Support Vector Machines (SVM), we observe that other learning algorithms scale  well. With Feed-Forward Neural Network (FFNN) having the greatest variability in its gains.  This is because FFNN is often stuck in local minima when learning starts. Random Forest (RF)  also steadily increases its test accuracy when the dataset is scaled. The Convolutional Neural  Networks (CNN), which never stuck in local minima, makes the highest gain.

\begin{figure}[h!]
	\centering
	\includegraphics[scale = 0.45]{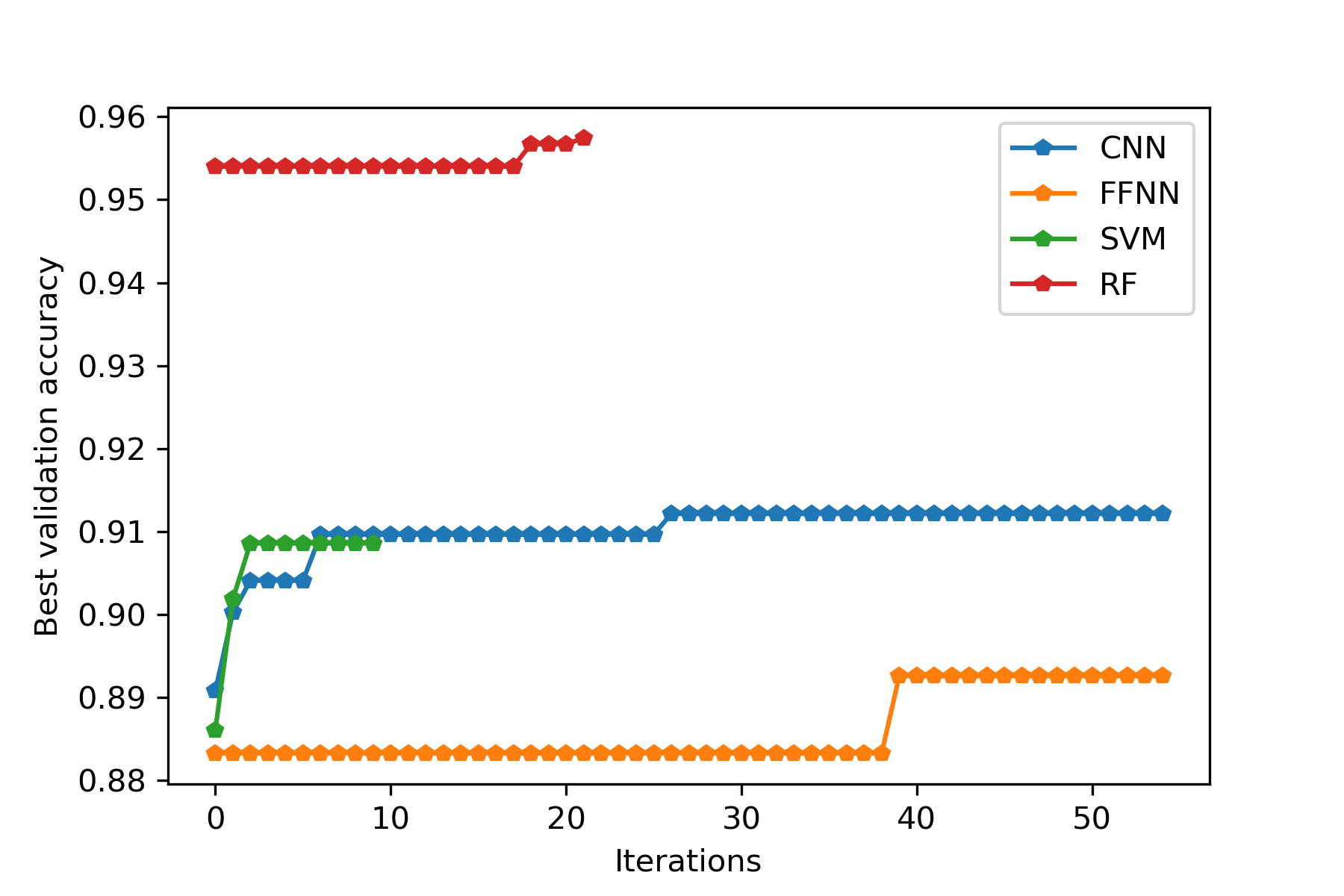}
	\caption{Illustration of the hyperparameters selection with Bayesian optimization procedure for Random Forests (RF), Support Vector Machines (SVM), Feed-Forward Neural Networks (FFNN) and Convolutional Neural Networks (CNN). \label{fig:optimization}}
\end{figure}

\begin{figure}[h!]
	\centering
	\includegraphics[scale = 0.45]{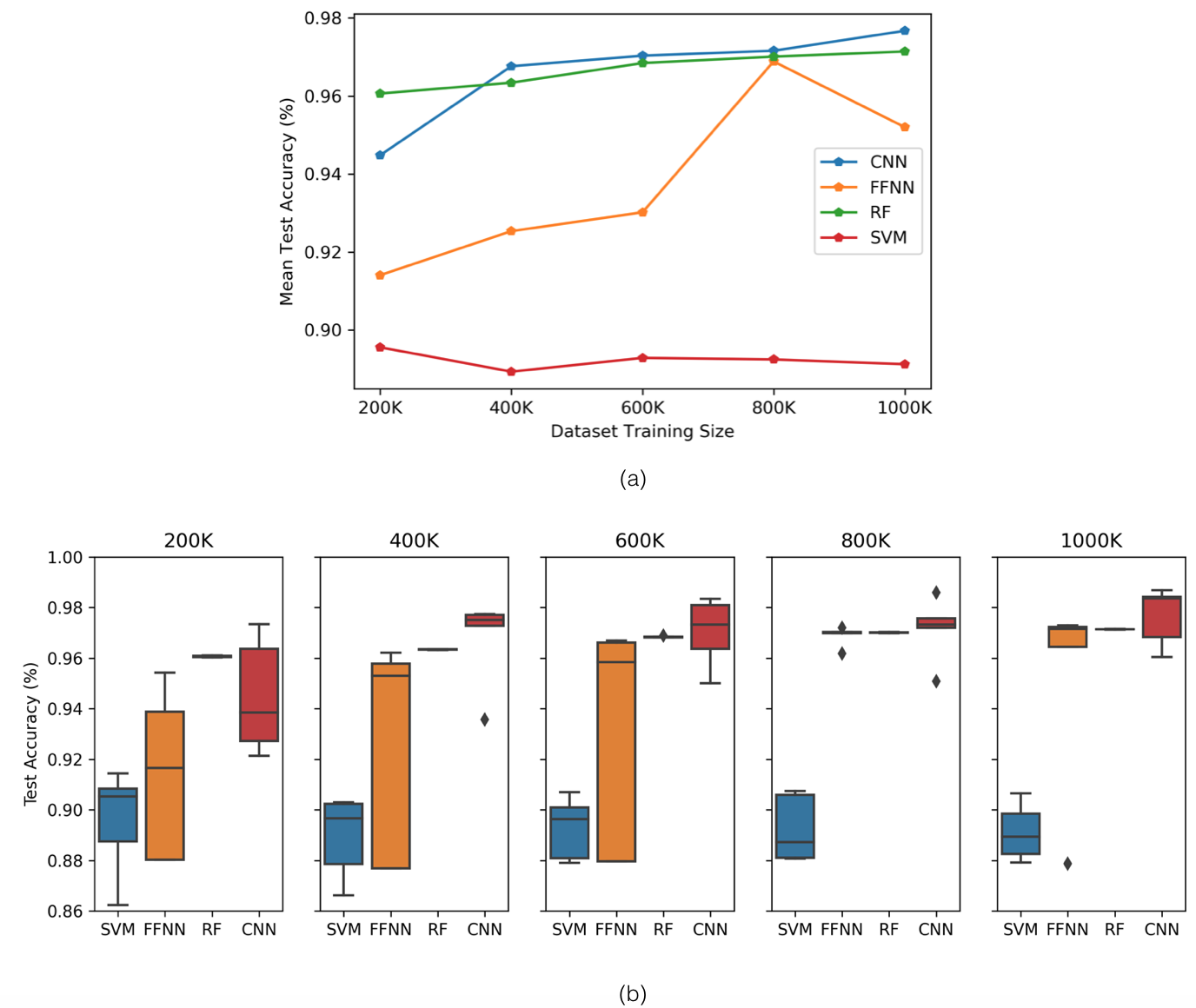}
	\caption{(a) Mean accuracy got for Random Forests (RF), Support Vector Machines (SVM), Feed-Forward Neural Networks (FFNN) and Convolutional Neural Networks (CNN) while varying the training dataset size from 200 000 to 1000 000 pings from the 2011 sea survey dataset. (b) Test accuracy summary statistics of each learning algorithm after 5 repetitions of the training process.}
\end{figure}

\begin{figure}[h!]
	\centering
	\includegraphics[scale = 0.45]{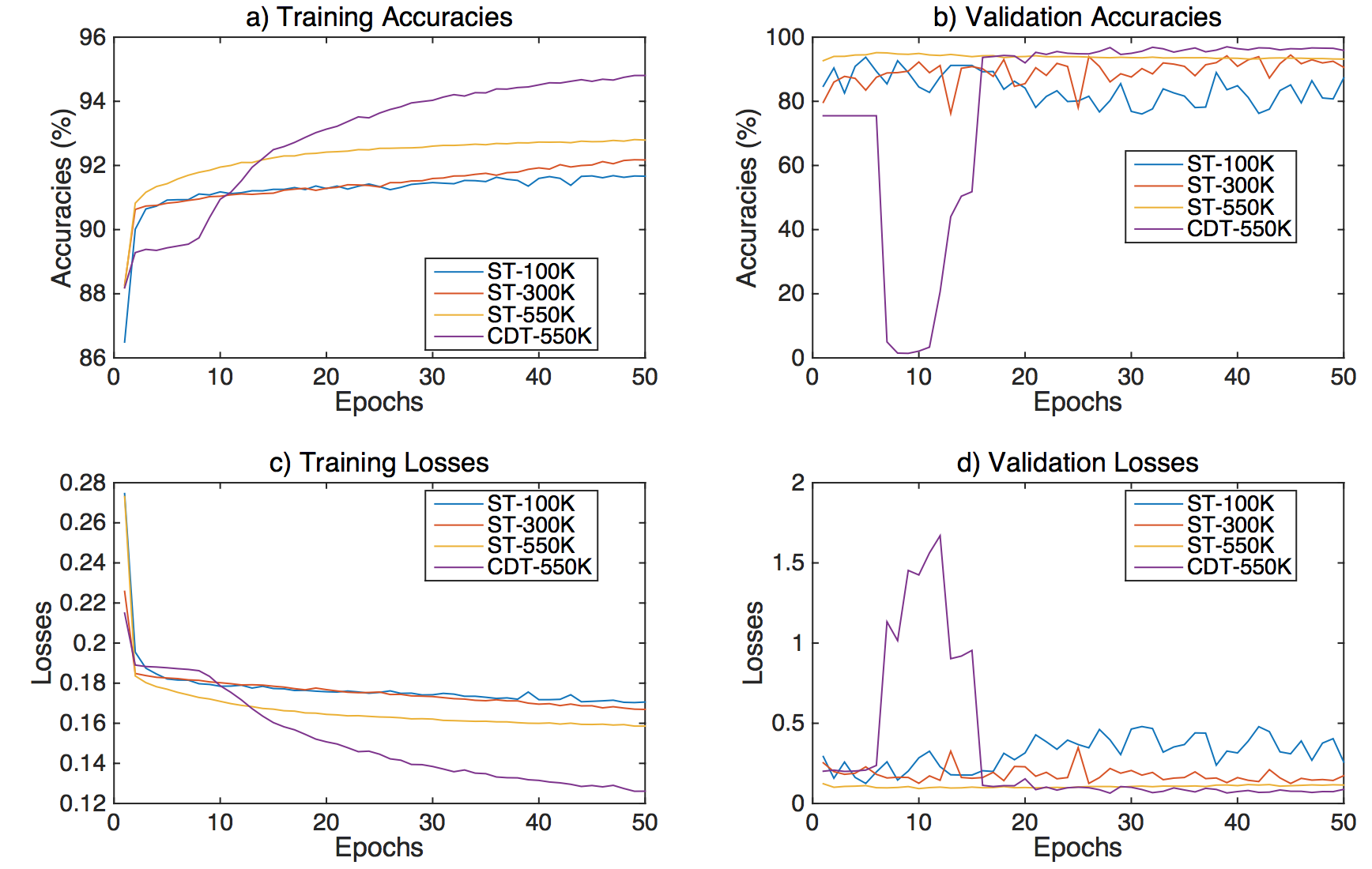}
	\caption{Performance at each iteration (epoch) of the model during training and validation on: simple training (ST) and cross-domain training (CDT) for 100,000, 300,000, and 550,000 pings. (a) training accuracy, (b) validation accuracy, (c) training losses and (d) validation losses. Validation accuracy and losses were obtained during training on a single unseen validation set from the 2015 sea survey dataset.}
\end{figure}

\begin{figure}[h!]
	\centering
	\includegraphics[scale = 0.45]{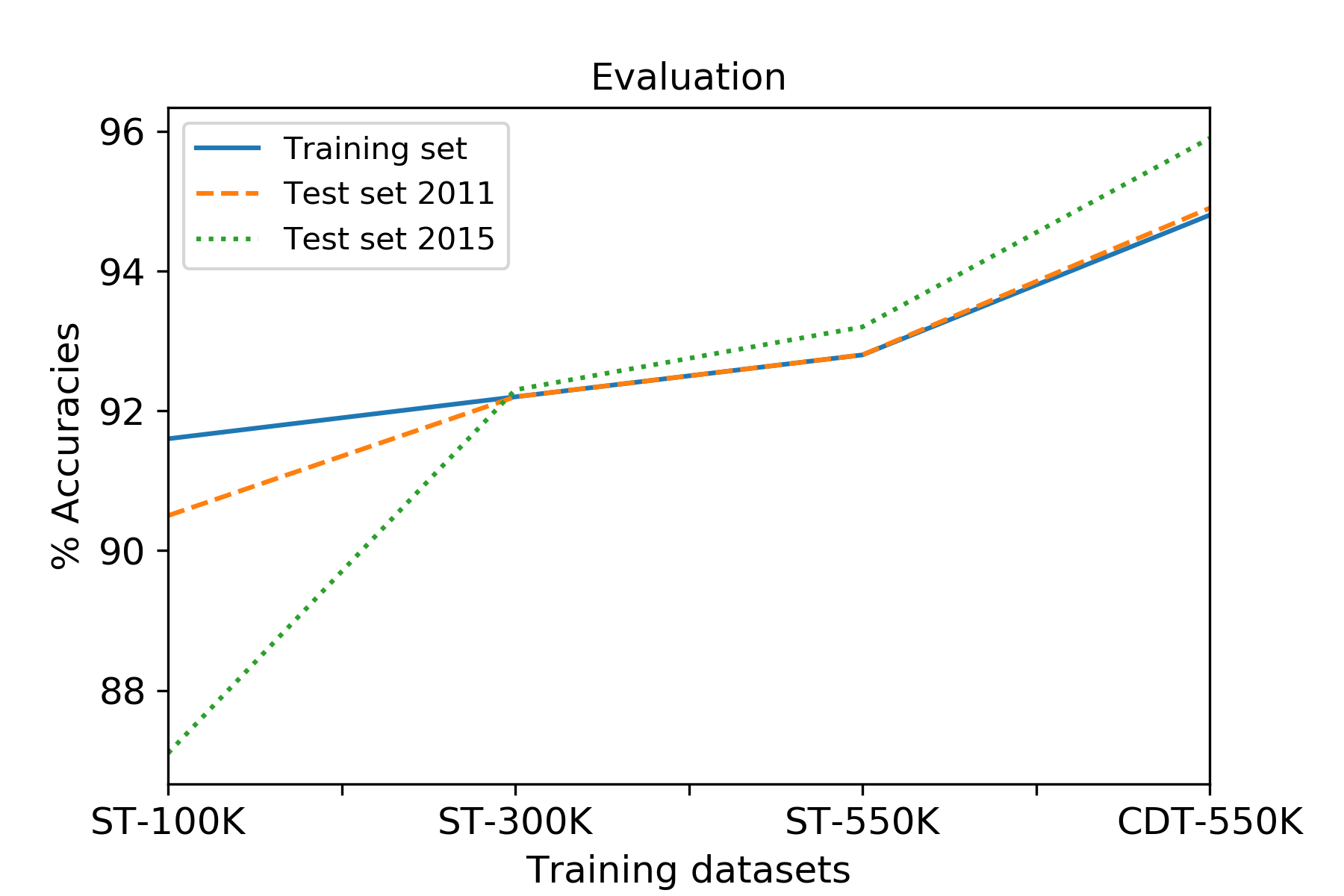}
	\caption{Evaluation of model accuracy on training dataset (solid line), 2011 test dataset (red circled line) and 2015 test dataset (green circled line) when increasing the training dataset size from 100,000 pings to 550,000 pings. See training datasets details in Section 2.1.4. . (For interpretation of the references to color in this figure legend, the reader is referred to the web version of this article.)}
\end{figure}

\subsection{Improving generalization accuracy with a cross domain dataset}  
\subsubsection{Learning evaluation on the training set }
Training accuracies showed little difference when training on ST-100K and ST-300K; the latter  displayed a slightly better evolution after epoch $\sim$25 (after 50 epochs respectively, 92\%, and  92\%; Fig 6a). With ST-550t, a net increase in training accuracy appeared after epoch 2 and  reached 93\% at the end of the training. When training on CDT-550K, the training accuracy  increased slowly until epoch $\sim$8, but overtook the other training experiments after $\sim$15 epochs  and finally reached 95\% (Fig 6a). The training losses displayed similar (but symmetric)  tendencies, reaching final losses of 0.17, 0.16, 0.15, and 0.12, respectively, for training with  100,000, 300,000, and 550,000 pings from the 2011 dataset, and 550,000 cross-domain pings  (Fig 6c). 

\subsubsection{Learning evaluation on the validation set }
In contrast to the continuous increase in training accuracy, in the case of simple training, the  validation accuracy displayed a higher variance, with a flat tendency among epochs; the  variability was smaller, however, for larger training datasets (Fig 6b). The average accuracy  reached 87\%, 91\%, and 93\%, respectively, for datasets ST-100K, ST-300K, and ST-550K at  epoch 50. Note also that the validation accuracy obtained when training on ST-100K decreased  over the epoch (Fig. 6b), which denoted a degree of over fitting. The effect of using a cross domain dataset for training was more evident from the validation performances. Indeed, the  accuracy dropped to low values (0.1\%) around epoch $\sim$10, but then rapidly reached higher values  than those of the simple training set experiments after epoch 18, and then displayed a steady  upward tendency, reaching a value of 96\% at the end of the training process. Validation losses  displayed the same change in variability among training experiments. The validation losses for  simple training were 0.25, 0.17, and 0.11, and the final value for cross-domain training was 0.08  (Fig 6d). In summary, training on a cross-validation dataset yielded unstable performances  during the first $\sim$15 epochs, however, after epoch 30, it outperformed training being done with  simple datasets.  

\subsubsection{Generalization Performances}
The generalization performance was evaluated by applying the CNN to predict ping classes on  an unseen test set extracted either from the 2011 dataset or from the 2015 dataset (Fig 7). On the  2011 test set, the CNN trained with the ST-100K, ST-300K, ST-550K, and CDT-550K training  sets correctly classified 90\%, 92\%, 93\%, and 95\% of the pings, respectively. On the 2015 test  set, they correctly classified 87\%, 92\%, 93\%, and 96\% of the pings, respectively. The CNN 

trained on only 100,000 pings from 2011 (ST-100K) over fitted, as its performance was better on  the training data than on unseen data. The ST-300K training set led to a similar performance on  all test sets. The ST-550K training set led to the same accuracy on the training set and on the  2011 test set. Unexpectedly, better accuracy was obtained when evaluating the 2015 test set than  on the training and 2011 test sets. Finally, training on CDT-550K led to a leap in performance  both on the training set and the 2011 test set but yielded an even better performance on the 2015  test set. 

\section{Discussion}
Machine learning algorithms have been widely used in passive acoustics. Indeed, Shamir et al. (2014) introduced an automated method for target classification in bioacoustics. Yue et al. (2017) used support vector machines (SVM) and convolutional neural networks (CNN) for  underwater target classification. Chi et al. (2019) used feed-forward neural networks (FFNN) for  source ranging. Hu et al. (2018) applied CNN for underwater acoustic signal extraction. Many  machine learning methods were employed for underwater source localization: Niu et al. (2017)  compared SVM, Random Forests (RF), and FFNN, then, Ferguson et al. (2018) applied CNN for  the same task and Wang et al. (2018) compared traditional match field processing (MFP) with  FFNN and generalized regression neural networks. Finally, deep learning methods were also  applied for underwater source localization: for instance, Huang et. al (2018) used FFNN in a  shallow water environment, moreover, Niu et al. (2019) used deep CNN in the context of big  data. 

Yet, very few authors have applied deep learning algorithms on fisheries (active)  acoustics data (see section 1.2; Williams 2016, Denos et al. 2017, Brautaset et al., 2020). A recent literature review of ML in acoustics is provided by (Bianco et al. 2019). One premise of  deep learning is that it could allow us to treat crude data in an end to end fashion (LeCun et al.  2015).  

Here we investigate ML methods on pre-processed echograms (see section 2.1.1). Using  Matecho (Perrot et al. 2018) allows a full chain of processing methods to extract information and  perform echo-integration on echosounder data following an international standard. The present  study contributes to filling this gap and shows the potential of learning algorithms to serve as a  useful tool for fisheries acoustics expert processing tasks. Below we first discuss the pipeline  experimented here for bottom correction in fisheries acoustic dataset. Then, we identify the  remaining challenge for this approach. We conclude on the main contribution of the paper and  provide perspectives one the potential benefit of using ML in fisheries acoustics.  
Two patterns emerge from our experiments of applying ML to active acoustics data, with  one of them fairly unexpected. Firstly, as expected in DL, increasing the size of the training data  almost always leads to better performance on the training set but also on each test set. Secondly,  using a cross-domain dataset for training leads to a leap in accuracy during training, validation,  and at test time. This emphasizes the sensitivity of the generalization performance to the  diversity of the training dataset. 

\begin{table}[h!]
\centering
\caption{Accuracies evaluated by each model on the 2011 sea survey dataset and the 2015 dataset. It seems that adding a subset of the 2015 data within the training set helps the model to get a better performance on both test set. In bold the upper and lower accuracies.}
    \begin{tabular}{|c|c|c|c|}
    \hline
    Final accuracies (in \%) & Training set & Test set — 2011 & Test set — 2015 \\
    \hline
    ST — 100k & 92 & 91 & 87 \\
    \hline
    ST — 300k & 92 & 92 & 92 \\
    \hline
    ST — 550k & 93 & 93 & 93 \\
    \hline
    CDT — 550k & 95 & 95 & 96 \\
    \hline
    \end{tabular}%
\end{table}

\begin{table}
\centering
\caption{Hyperparameters search space and value found using a Python Bayesian optimization library (GyOpt).}
\begin{subtable}[h]{0.6\textwidth}
    \centering
    \begin{tabular}{|p{6cm}|p{2cm}|p{2cm}|}
    \hline
    \multicolumn{3}{|c|}{Random forest} \\
    \hline
    Hyperparameter & Search Space & Value\\
    \hline
    Number of tree range & [10, 10 000] & 187 \\
    \hline
    Min. samples leaf & [20, 50] & 24 \\
    \hline
    \end{tabular}%
\end{subtable}
\begin{subtable}[h]{0.6\textwidth}
    \centering
    \begin{tabular}{|p{6cm}|p{2cm}|p{2cm}|}
    \hline
    \multicolumn{3}{|c|}{Support vector machines} \\
    \hline
    Hyperparameter & Search Space & Value\\
    \hline
    Alpha & [0.0001, 0.1] & 0.077 \\
    \hline
    \end{tabular}%
\end{subtable}
\begin{subtable}[h]{0.6\textwidth}
    \centering
    \begin{tabular}{|p{6cm}|p{2cm}|p{2cm}|}
    \hline
    \multicolumn{3}{|c|}{Feed-forward neural network} \\
    \hline
    Hyperparameter & Search Space & Value\\
    \hline
    Number of neuron: fully connected layer 1 & [5, 600] & 75 \\
    \hline
    Number of neuron: fully connected layer 2 & [5, 320] & 105 \\
    \hline
    Number of neuron: fully connected layer 3 & [5, 120] & 95 \\
    \hline
    Dropout rate: fully connected layer 3 & [0, 1] & 0.6 \\
    \hline
    \end{tabular}%
\end{subtable}
\begin{subtable}[h]{0.6\textwidth}
    \centering
    \begin{tabular}{|p{6cm}|p{2cm}|p{2cm}|}
    \hline
    \multicolumn{3}{|c|}{Convolutional neural network} \\
    \hline
    Hyperparameter & Search Space & Value\\
    \hline
    Kernel 1 & [5, 60] & 5 \\
    \hline
    Kernel 2 & [5, 60] & 59 \\
    \hline
    Kernel 3 & [5, 60] & 19 \\
    \hline
    Number of neuron: fully connected layer 1 & [5, 600] & 260 \\
    \hline
    Number of neuron: fully connected layer 2 & [5, 320] & 319 \\
    \hline
    Number of neuron: fully connected layer 3 & [5, 120] & 101 \\
    \hline
    Dropout rate: fully connected layer 3 & [0, 1] & 0.9 \\
    \hline    
    \end{tabular}%
\end{subtable}
\end{table}

\begin{figure}[h!]
	\centering
	\includegraphics[scale = 0.75]{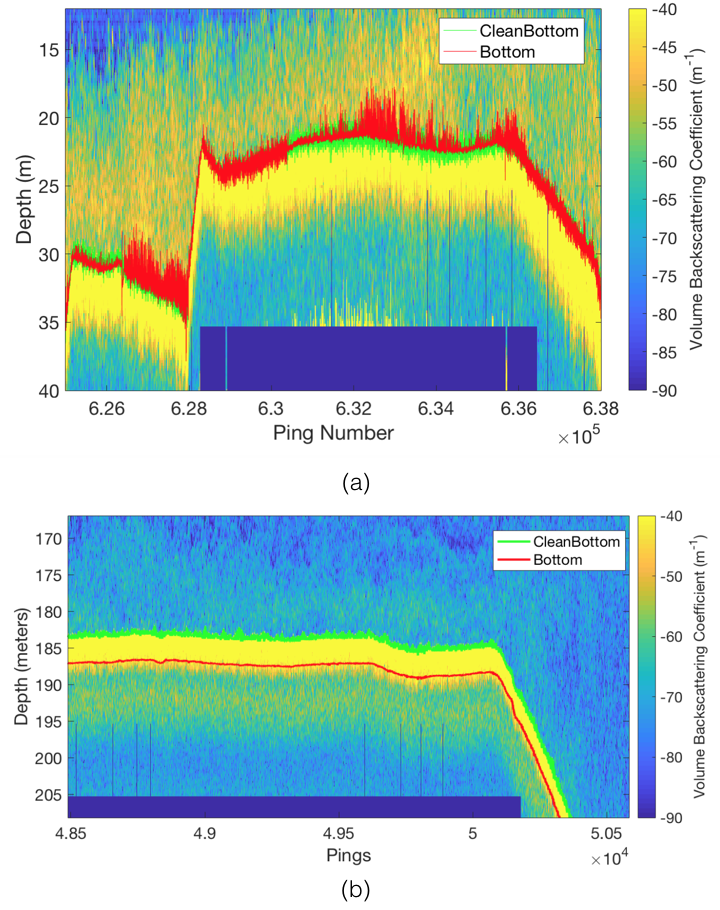}
	\caption{(a) Example of echogram in which automatic bottom detection (red line) failed due to presence of a planktonic layer directly above the sea floor (soft yellow). The green line is the bottom line after correction by the expert. (b) Example of echogram in which automatic bottom correction (red line) failed due to the transducer depth being moved down during the sea campaign showing a constant offset shift. . (For interpretation of the references to color in this figure legend, the reader is referred to the web version of this article.)}
\end{figure}

\subsection{Learning algorithm comparison: why CNN stand out }
SVM performed poorly, indeed this is because we used a linear kernel on a high dimensional  problem, this was required to allow the SVM to learn on large datasets. Also as SVM do not  allow to increase the number of parameters used to learn, it has a lower asymptote than the other  models (Fig 5.a). While RF has the best accuracy when trained with a small dataset (200K), CNN proved to have better performance as soon as the datasets got bigger. Yet, RF is the second  best option and followed the CNN closely even though the accuracy obtained with 1000K pings  seemed to led the CNN accuracy to take off. Another interesting property of RF is that they  exhibit almost no variability when compared among different training versions on the same  dataset. This property allows the modeler to be sure that he got the best possible achievable  result with a RF when training is achieved. The main advantage of neural networks, whether  FFNN or CNN, is the ability they gave to the modeler to increase the number of the parameters  to the learning problem. As a result, they are modular and can adapt to many kinds of problems.  The main downside of FFNN was its variability from training to training. Indeed the reader can  observe that the worst result obtained for each dataset is closely the same around 87\% (Fig 5.b).  At least for one run over the 5 we did, the FFNN achieved almost no learning. This can happen  when the network is stuck in a local minimum and cannot get outside of it during training. In  contrast, during training, the CNN has never fallen in a local minimum (Fig 5.b). This is because  the convolutional layers compressed the original data in a lower-dimensional representation that  is further passed to a fully connected neural network. As a result, the important features are  summarised by the convolutional layers (Goodfellow et al. 2016, Lecun et al. 2015). And the  optimization is made in a lower-dimensional space. Convolution has shown great promise for  image understanding. The above comparison suggests that 1-dimensional convolution is the  more adapted class of learning algorithm to process active echosounder data. Finally, CNN  exhibits the best performance with training datasets of sizes 400K, up to 1000K. Niu et al.  (2017) found comparable performance for each model: SVM, RF, and FFNN in underwater  source localization framed as a classification problem. Our results differed from those of Niu et  al. (2017) as we found a clear advantage of using neural networks and more specifically convolutional neural networks. This difference can be explained as they used a relatively small  dataset for training (1,380 samples for training, 120 examples for testing), also their work was on  simulated passive acoustic data. In contrast, our work was done on a real-world and bigger  dataset (200,000 examples to 1,000,000 examples) of active acoustic data. Getting better results  on bigger datasets with Deep Learning is discussed in Goodfellow et al. (2016) and Sun et al. (2017). Furthermore practical applications were found in underwater source localization by Niu  et al. (2019) who used a 50 layers residual network to learn on tens of millions of training  examples. 

\subsection{Cross domain training as a way to tackle imbalances and irregularities in datasets}
The specific problem we tacked is the task of helping the expert to label a new dataset faster  given past labeled data. An important assumption in ML is that the data we want to learn from as  well as the future unseen data must come from the same distribution and be identically and  independently distributed (Goodfellow et al. 2016). This is the first challenge that rose from the  task at hand. This assumption does not hold in our case, as discussed in section 2.1.3 for the label  distribution and as illustrated in Fig 2 for the noise distribution.  

Methods that deal with unbalanced datasets exist (Shimodaira 2000; Crammer et al.  2008). However, these techniques need to know the distribution of the target dataset to correct  imbalances during training. So, they do not apply in our case because the label distribution of a  new dataset is unknown before the expert has labeled it. One could say that the expert could label  a sample of the new dataset to get an estimation of the target label data distribution. There are  approaches in the ML literature that deal exactly with this problem for instance semi-supervised  learning (Chapelle et al. 2006) and active learning (Settles 2010). While these approaches could be envisioned in future work, they need to take into account the specificity of fisheries acoustics  data. Indeed the labeling needs to occur for several pings following each other because the expert  needs to get context from the ping in the same area to accurately label the bottom line. In a sense,  the labeling process is not based on the review of individual pings but on the review of chunks of  pings. This contrasts with traditional computer vision tasks, where the labeling of each image is  independent of each other. In that sense, traditional active learning methods could be useful if  applied to a chunk of data, or to individual pings with the goal to review every ping in a given  neighborhood. Similarly, semi-supervised learning methods require training a model on a  representative sample of labeled data, augmented with unlabeled data. The rationale is to use the  structure of unlabeled data to learn useful features of the model that can be refined on the labeled  dataset during training. This condition could be met if the expert selects a representative subset  of ping and labels their neighborhood. In this paper, we took the simplest possible setting: to  label a randomly selected chunk of 100,000 pings from the new dataset. As this is relatively easy  to do in practice. Indeed, this chunk of data represents $\sim$5\% of the total number of pings.  Furthermore, we quantified if this chunk of pings could benefit the model at training time by  monitoring its performance on a validation set made of this chunk of data (Fig 6). Finally, we  evaluated the model trained in each setting, on two large test sets made of respectively all unseen  ping from the 2011 and the 2015 pool. We observe that using a cross-domain dataset improves  the performance on each test set (Fig 7), suggesting that novel features introduced by the new  dataset add to the complexity of the cross-domain training set as opposed to the simple training  set. Indeed, even the accuracy on the training set improved. The optimization procedure during  training seems to be less stable in the first few epochs, this is probably due to a different path  taken by the algorithm when updating its weights. 

Also, note that while the current system got 96\% accuracy (Table III) on the 2015 test set,  its performance could be enhanced by training on a larger database composed of diverse sea  surveys. Indeed, we have shown that training with an increasingly bigger training set size leads  to better results with local test sets (Fig 3; Fig 5). In addition, we have shown the potential of  mixing datasets (Fig 7). This is in line with the results of Sun et al. (2017), more data is always  welcome for deep learning models.

\subsection{Challenges of Machine Learning methods with Fisheries Acoustics Data }
Although the 2011 and 2015 datasets were collected in the same area, i.e., the northwest African  continental shelf, using the same vessel, there was a great divergence in the error distribution of  the initial bottom line estimation, which was subsequently corrected by experts. There were also  irregularities in the noise distribution of the data between the 2011 and 2015 echograms (Fig 2).  Indeed, it is often the case that datasets collected from different sea surveys present different  attributes even if they were collected in the same geographical area. This may be due, for  example, to different weather conditions encountered during the two sea surveys (wind and sea  agitation). Indeed, the bottom reflection signal can be altered by air bubbles (generated by wave  breaking), as well as by the ship's roll and pitch. Bottom line correction also depends on the  biological resources targeted. For example, the expert corrects the bottom line more carefully  when the echointegration concerns resources close to the bottom line, as opposed to pelagic  species occurring in the water column, since a precise bottom estimation is less crucial for the  latter. As a result, even during post-processing by the expert for bottom correction, we can have  part of the echogram being corrected with less accuracy. Errors can be caused by the detection of  biological resources in contact with the bottom (e.g., planktonic layer or fish school) (Fig 8.a) or a change in the transducer depth settings by an operator on board the vessel (Fig 8.b). Also,  different settings of the echosounder during the data collection may yield a different noise  distribution. These differences between different datasets may explain the surprising result of  better performances on the 2011 test set than on the training set in the case CDT-550K (Fig 7).  However, it does not clearly explain why the CNN trained with a CDT-550K performed better at  classifying the 2015 test set than classifying the 2011 test set (Fig 5).  

One hypothesis that may explain this phenomenon is that the 2011 dataset benefited from  a much greater degree of human correction than the 2015 dataset. Indeed, in proportion, more  pings have been corrected in the 2011 dataset than in the 2015 dataset as discussed in section  2.1.3. Consequently, we learned on a dataset with a higher likelihood of including bad labeling,  in other words, we trained on a dataset harder to learn. Yet, as stated in his survey of DL  methods in remote sensing (Ball et al. 2017), the question of how the DL algorithm ingests non 
heterogeneous data sources is still open. Our interpretation, though, is that since the loss function  is not convex, the minimization problem admits multiple local minima, thus ingesting  heterogeneous data may influence the weights update trajectory during training. And a more  diverse training set leads to better minima.  

\subsection{Limitations due to the labeling process }
The other obvious limitation is relative to the data labeling process. We discriminated our data  based on a threshold that we found when computing the numerical difference between the  bottom line calculated by the onboard procedure and the expert post-processing corrected bottom  line (referred to as “clean bottom”). However, in some cases this difference was not due to  echogram noise that could be detected by an algorithm (as shown in Fig 8.a). For example it could occur if the transducer was moved up or down (on the onboard setting or concretely)  during the sea campaign, as a result the bottom given by the procedure is under the real bottom,  even if the bottom line appears from the echogram structure (Fig 8.b). The issue is that such  echograms are labeled as needing expert correction while nothing should have generated an error  in bottom localization, since the error comes from an external reason (here a change of the  transducer depth settings during the sea campaign). Thus, this kind of labeling in the training  dataset limits the quality of the learning process. 

Finally, we have reported several factors that could influence the quality of the training dataset and limit the classification performance that can be achieved with a trained algorithm on  different dataset. Indeed, the occurrence of cases where the bottom correction errors do not come  from the echogram structure alone might prevent an optimal classification by the trained CNN. 

\subsection{Perspective and Future Works }
We see two main dimensions in which this work can be extended: (1) by improving the  efficiency and the quality of the methodology, (2) by working toward an operational tool for the  fisheries acoustics community. 
A more careful annotation of the data would yield better results in terms of accuracy and  interpretability of how the model learns. In this respect, changing the data labeling procedure can  bring significant improvements. For example, instead of using a criterion based on the difference  in the distance between the bottom and the clean bottom, one could compare the amount of  energy in the echogram (in mean volume backscattering strength (Sv in dB)) at the respective  depths of the bottom and the clean bottom. In the same way, it might be interesting to use a depth  offset for a given time interval, with the rationale that the bottom value cannot jump, given two consecutive pings, except when there is a particular bottom relief. Also, a multi-criteria  procedure for labeling each ping could be considered if the computing power available for  training allows it. Future studies will need to experiment with more datasets from diverse  geographic areas to further study the benefits of generalization from learning with multiple  datasets. In particular, this would allow further study on the effects of data irregularities and  distribution differences in the learning process. A good starting point toward that goal would be  to use active learning or semi-supervised learning methodologies (Chapelle et al. 2006; Settles  2010). 

We advocate that working upfront on a unified data collection procedure would certainly  reduce errors due to data irregularities and also reduce the variability coming from different  experts concerning the labeling process. Using machine learning to standardize the labeling  process across geographical regions would be useful to enable more meaningful application for  the downstream modeling task of fish density estimation. A step toward that direction would be  to create a common pool of labeled fisheries acoustics data and to train a deep convolutional  neural network on it. This would result in the best performance achievable because as we have  shown, the quantity and the diversity of the dataset play a key role to improve the performance of  DL models. The trained model could then be used to standardize the labeling of new sea surveys  collected globally. One of the benefits of having a uniform labeling process across geographical  regions is to allow for comparative studies of new computational methods in fisheries acoustics.  

\section{Conclusion}
The goal of this work was to develop a system designed to help a human expert to correct the  bottom line from echogram data. Our main contribution to the study of fisheries acoustics has been to establish that CNN are able to extract useful features from different underwater active  acoustic data sources. To address inherent imbalances in data distribution, as well as  irregularities, a good practice is to learn with a cross-domain training set. The model trained can  be used to help the fisheries acoustics expert to find pings that are likely to require correction on  the bottom line with a high accuracy (>95\%). However, our results suggest that much better  classification accuracy can be reached using larger and more diverse datasets for training; such  an approach could easily be implemented using dedicated GPUs. Finally, this work demonstrates  the potential of CNN to learn features from fisheries acoustics data. This suggests that DL could  be used furthermore to standardize the labeling process across regions and open the road to interesting areas of research based on large historical acoustic datasets in which other objects  such as fish schools have been labeled (e.g., Scalabrin and Massé 1993; MacLennan et. al. 2004;  Trygonis et al. 2016) to better discriminate between fish schools and sound scattering layers  (Diogoul et al. 2020). Particularly demersal fish schools near the bottom as well as seabed  echoes.  

\section{Acknowledgments}
We are grateful to the AWA Project (“Ecosystem Approach to the management of fisheries and marine environment (EAMME) in West African waters”) funded by the German Federal Ministry of Education and Research (BMBF) and the French Research Institute for Development (IRD) (Grant 01DG12073E) implemented by the Sub-Regional Fisheries Commission (SRFC), the Preface Project funded by the European Commission’s Seventh Framework Program (2007–2013) under Grant Agreement No. 603521, EU TriAtlas project (Grant Agreement No. 817578) and the FAO/Nansen Project for data collection. We are grateful to Adrien Berne (IRD, France) who performed the manual clean-bottom corrections for both surveys, to Dr Jens-Otto Krakstad of the Norwegian Institute of Marine Research (IMR) who led the surveys at sea off northwest Africa, and to Gildas Roudaut and Jérémie Habasque (IRD, France) for their early interest in this approach and expertise in fisheries acoustics labeling. We are also grateful to Théophile Bayet (IRD) for constructive discussions and comments.

\bibliographystyle{apalike}  
\bibliography{references}
\nocite{*}

\end{document}